\documentclass[fleqn,table]{wlscirep}
\usepackage{color}
\usepackage{array}
\usepackage{amssymb}
\usepackage{afterpage}
\usepackage{balance}
\usepackage{tikz}
\usepackage{eso-pic}

\usepackage{amsmath}

\usepackage{floatrow}
\usepackage{enumitem}
\usepackage{multicol}

\usepackage[author={Manfred Eppe}]{pdfcomment}

\newcommand{\concept}[1]{\emph{#1}}

\usepackage{lipsum}

\usepackage{booktabs} 
\usepackage{siunitx} 
\usepackage{pgfplotstable} 

\usepackage[breakable, theorems, skins]{tcolorbox}
\tcbset{enhanced}

\usepackage{scrextend}
\changefontsizes[10pt]{9pt}
\usepackage{pgf}
\pgfdeclareradialshading{shadednodered}
{\pgfpoint{-0.2cm}{0.5cm}}%
{rgb(0cm)=(1,1,1);
rgb(0.9cm)=(0.9,0.3,0.1); rgb(1.4cm)=(0.7,0.2,0); rgb(4.4cm)=(0,0,0)}

\pgfdeclareradialshading{shadednodeorange}
{\pgfpoint{-0.2cm}{0.5cm}}%
{rgb(0cm)=(1,1,1);
rgb(0.9cm)=(0.95,0.5,0.2); rgb(1.4cm)=(0.7,0.35,0.1); rgb(4.4cm)=(0,0,0)}

\pgfdeclareradialshading{shadednodeyellow}
{\pgfpoint{-0.2cm}{0.5cm}}%
{rgb(0cm)=(1,1,1);
rgb(0.9cm)=(0.95,0.8,0.5); rgb(1.4cm)=(0.7,0.6,0.3); rgb(4.4cm)=(0,0,0)}

\pgfdeclareradialshading{shadednodewhite}
{\pgfpoint{-0.2cm}{0.5cm}}%
{rgb(0cm)=(1,1,1);rgb(4.4cm)=(1,1,1)}

\tikzset{
  LabelStyle/.style = { rectangle, rounded corners, 
                        draw,
                        fill = gray!50, 
                        font = \small},
TextStyle/.style = { font = \small},                        
  VertexStyle/.append style = { inner sep=5pt,
                                font = \small\color{black},
                                shape = rectangle, 
                                rounded corners,
                                minimum height = 1.5cm,
                                color=white,
                                shading = shadednodered,
                                align = center},
  EdgeStyle/.append style = {->, align = center} }
  
\usepackage{array}
\newcolumntype{L}[1]{>{\raggedright\let\newline\\\arraybackslash\hspace{0pt}}m{#1}}
\newcolumntype{C}[1]{>{\centering\let\newline\\\arraybackslash\hspace{0pt}}m{#1}}
\newcolumntype{R}[1]{>{\raggedleft\let\newline\\\arraybackslash\hspace{0pt}}m{#1}}
\usepackage{multirow}
\usepackage{colortbl}
\usepackage{makecell}

\definecolor{TableRowColor}{rgb}{0.9821,0.953,0.906}
\definecolor{TableHeadColor}{rgb}{0.9843,0.92157,0.82745}

\DeclareCaptionLabelSeparator{vertpipe}{ | }                
\usepackage{float}

\usepackage{titlesec}

\titleformat{\section}
  {\color{color1}\large\sffamily\bfseries}
  {}
  {0.0em}
  {#1}
  []

\titleformat{\subsection}
  {\sffamily\bfseries}
  {}
  {0.0em}
  {#1}
  []

\newcounter{infobox}

\usepackage[framemethod=tikz,rightmargin=0,
leftmargin=0,backgroundcolor=TableRowColor,
frametitlerule=false,frametitlebackgroundcolor=TableHeadColor,roundcorner=0pt,innerleftmargin=2pt,hidealllines=true]
{mdframed}

\newenvironment{infobox}[1][]{%
  \refstepcounter{infobox}
  \begin{mdframedtoc}[%
    frametitle={\footnotesize\sffamily Box \theinfobox\ | #1},
    ]%
  \small
  }{
  
  \end{mdframedtoc}
}

\makeatletter
\newmdenv[startinnercode=\addcontentsline{mdbox}{subsection}\mdf@frametitle]
{mdframedtoc}
\newcommand{\listofboxes}{%
 \section*{Liste des focus}
 \@starttoc{mdbox}%
}
\makeatother

\usepackage{nameref}

\makeatletter
\newcommand{\firstof}[1]{\@car#1\@nil}
\newcommand{\restof}[1]{\expandafter\@cdr#1\@nil}
\makeatother

\usepackage[textwidth=15mm]{todonotes}
\usepackage{marginnote}

\definecolor{revcolor}{RGB}{0,45,78}
\definecolor{verylightgray}{RGB}{240,240,240}

\newcommand{\rev}[2]{\todo[size=\scriptsize,backgroundcolor=verylightgray,bordercolor=verylightgray,linecolor=verylightgray,fancyline]{#2}{\color{revcolor}#1}}

\usepackage[style=nature,isbn=false,date=year,defernumbers=true,sorting=none,intitle=true,doi=true]{biblatex}

\DeclareSourcemap{
  \maps[datatype=bibtex]{
   \map{
      \step[typesource=techreport, typetarget=report]
    }
    \map{
      \step[fieldsource=pages, final]
      \step[fieldset=url, null] 
      \step[fieldset=doi, null] 
    }
    \map{
      \step[fieldsource=doi, final]
      \step[fieldset=url, null] 
    }
    \map{
      \step[fieldsource=eprint, null]
    }
  }
}
\DeclareFieldFormat{url}{Preprint at \url{#1}}
\makeatletter
\AtEveryCitekey{%
  \ifcsundef{blx@entry@refsegment@\the\c@refsection @\thefield{entrykey}}
    {\csnumgdef{blx@entry@refsegment@\the\c@refsection @\thefield{entrykey}}{\the\c@refsegment}}
    {}}
\defbibcheck{onlynew}{%
  \ifnumless{0\csuse{blx@entry@refsegment@\the\c@refsection @\thefield{entrykey}}}{\the\c@refsegment}
    {\skipentry}
    {}}
\makeatother

\newcommand{\citet}[1]{\citeauthor*{#1}\supercite{#1}}
\let\cite=\supercite
 \AtBeginBibliography{\small}

\renewcommand{\rev}[2]{#1}

\fancypagestyle{empty}{
  \lhead{This version of the article has been accepted for publication, after peer review and is subject to Springer Nature’s AM terms of use, but is not the Version of Record and does not reflect post-acceptance improvements, or any corrections. The Version of Record is available online at: \url{https://doi.org/10.1038/s42256-021-00433-9}}
}

\title{Intelligent problem-solving as integrated hierarchical reinforcement learning}

\author[1,*]{Manfred Eppe}
\author[2,3]{Christian Gumbsch}
\author[1]{Matthias Kerzel}
\author[1]{Phuong D. H. Nguyen}
\author[2]{Martin V. Butz}
\author[1]{Stefan Wermter}
\affil[1]{Universität Hamburg, Germany}
\affil[2]{University of Tübingen, Germany}
\affil[3]{Max Planck Institute for Intelligent Systems, Tübingen, Germany}
\affil[*]{manfred.eppe@uni-hamburg.de}

\begin{abstract}
According to \rev{cognitive psychology}{R3-\ref{rev:r3:final_edit}} and related disciplines, the development of complex problem-solving behaviour in biological agents depends on hierarchical cognitive mechanisms.
\rev{Hierarchical reinforcement learning is a promising computational approach that may eventually yield comparable problem-solving behaviour in artificial agents and robots. 
However, to date the problem-solving abilities of many human and non-human animals are clearly superior to those of artificial systems.
\rev{Here, we propose steps to integrate biologically inspired hierarchical mechanisms to enable advanced problem-solving skills in artificial agents.}{R3-\ref{rev:r3:clarify_contrib}}
Therefore, we first review the literature in \rev{cognitive psychology}{R3-\ref{rev:r3:final_edit}} to highlight the importance of  compositional abstraction and predictive processing.
Then we relate the gained insights with contemporary hierarchical reinforcement learning methods.
Interestingly, our results suggest that all identified cognitive mechanisms have been implemented individually in isolated computational architectures, raising the question of why there exists no single unifying architecture that integrates them.
As our final contribution, we address this question by providing an integrative perspective on the computational challenges to develop such a unifying architecture.
We expect our results to guide the development of more sophisticated cognitively inspired hierarchical machine learning architectures.}{E-\ref{rev:editor:abstract}}
\end{abstract}

\makeatletter
\let\newtitle\@title
\let\newauthor\@author
\let\newdate\@date
\makeatother

\begin{document}
\newrefsegment
\rfoot{\small\sffamily\bfseries\thepage/9}%
\twocolumn

\flushbottom
\maketitle

\thispagestyle{empty}



\begin{figure*}[ht]
    \centering
    \includegraphics[width=.19\columnwidth]{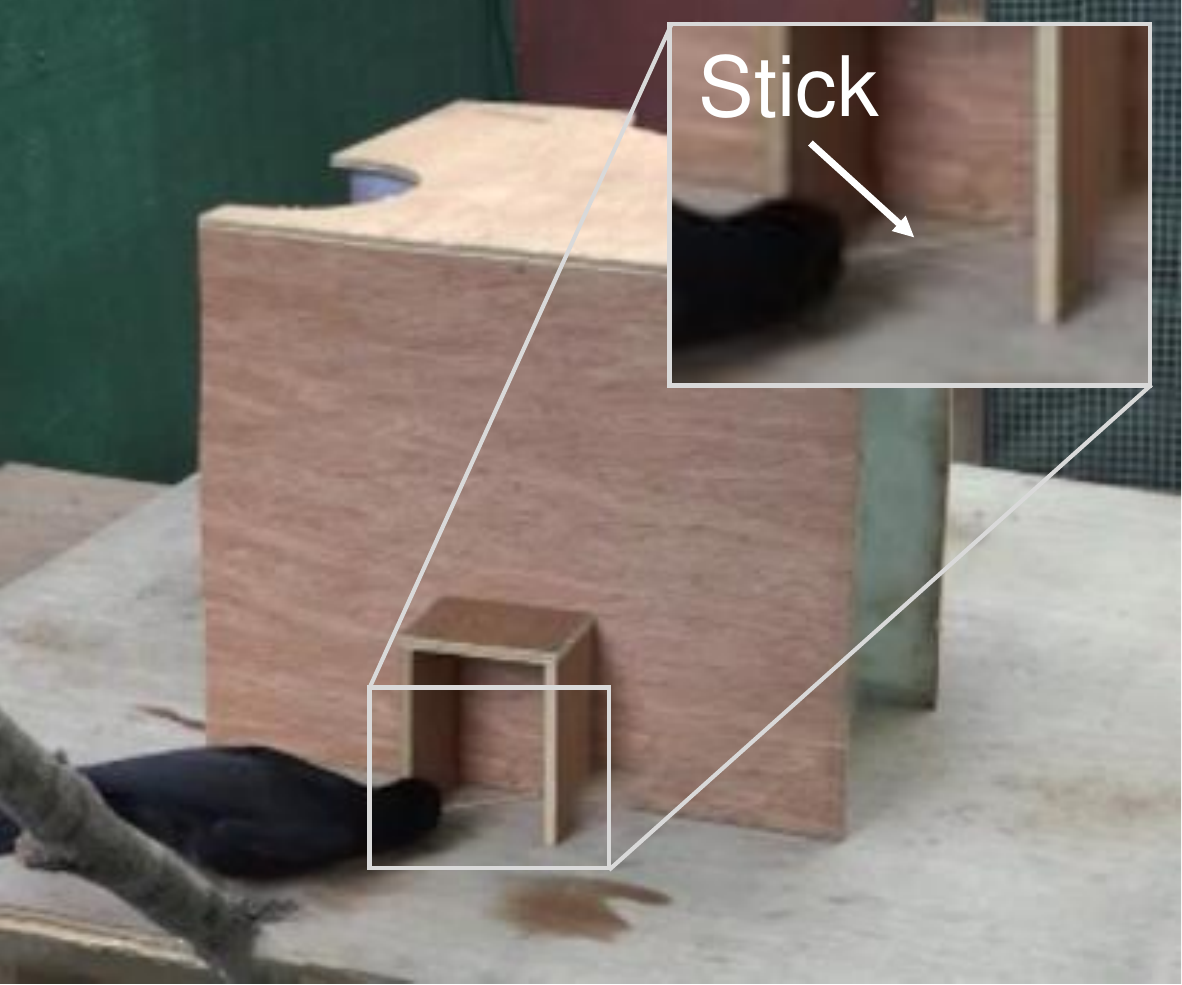}
    \hspace{-0.7cm}\begin{minipage}[b][0.3cm][t]{0.3cm}\textbf{(a.)}\end{minipage} 
    \hspace{0.3cm}
    \includegraphics[width=.19\columnwidth]{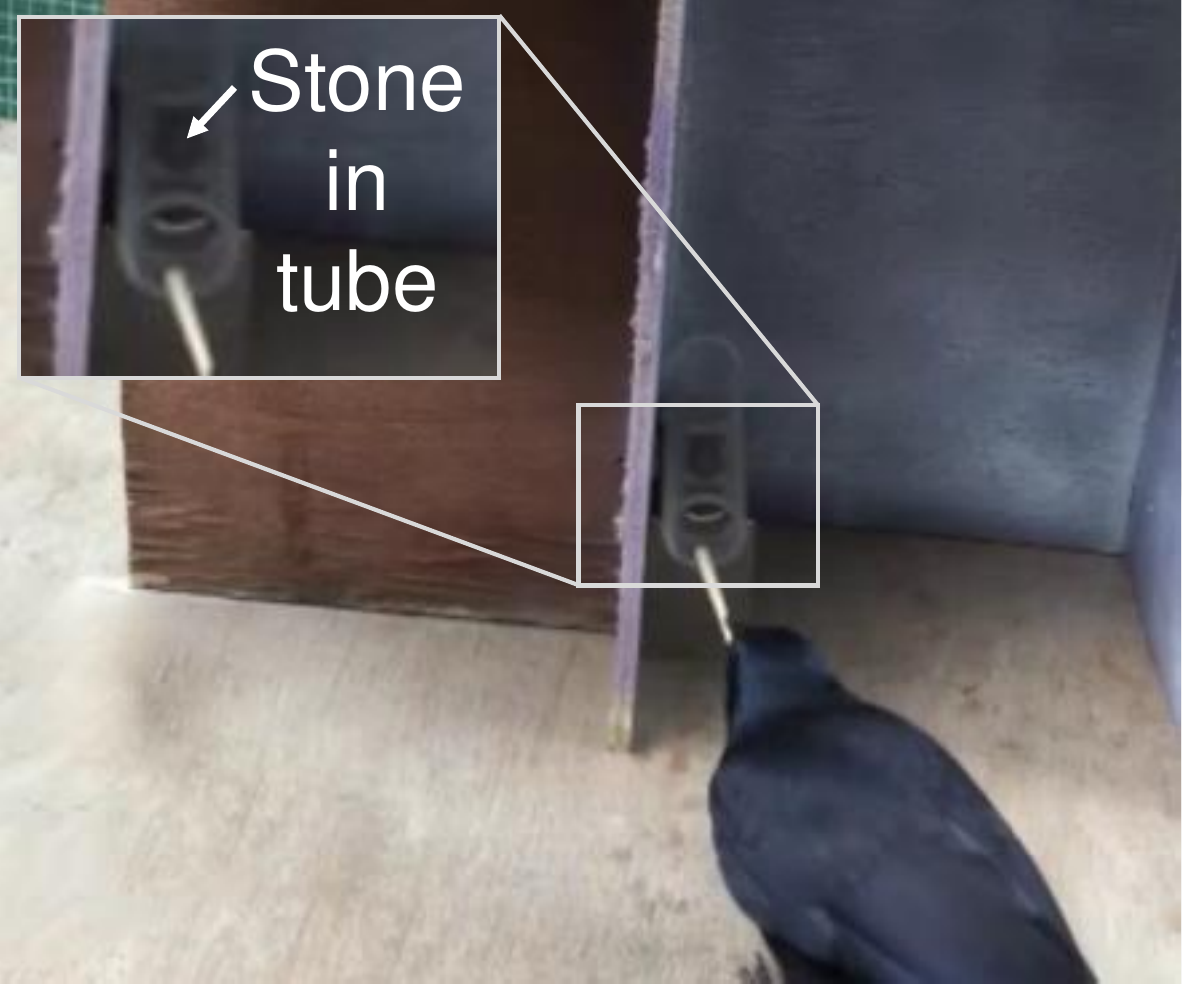}
    \hspace{-0.7cm}\begin{minipage}[b][0.3cm][t]{0.3cm}\textbf{(b.)}\end{minipage} 
    \hspace{0.3cm}
    \includegraphics[width=.19\columnwidth]{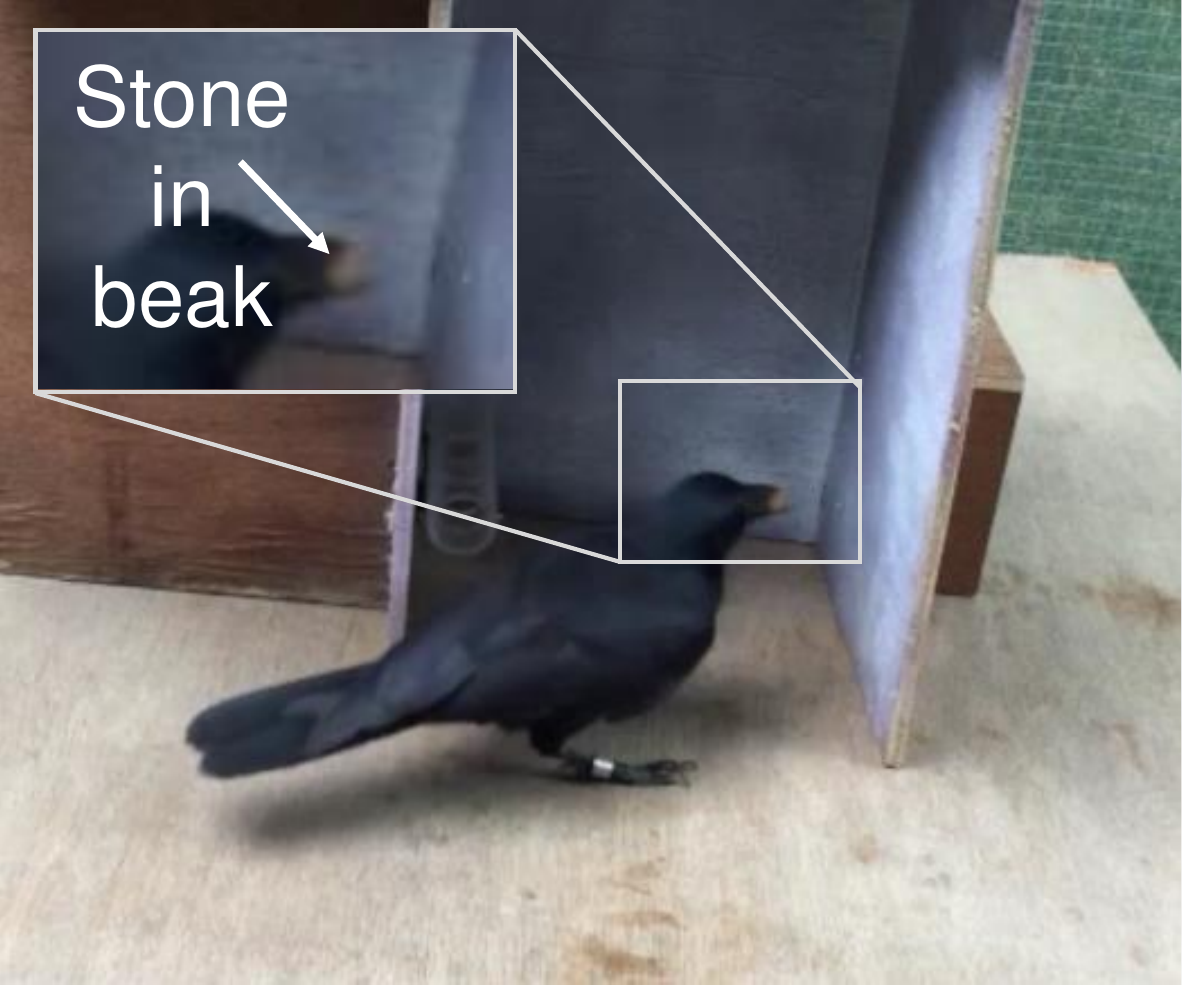}
    \hspace{-0.7cm}\begin{minipage}[b][0.3cm][t]{0.3cm}\textbf{(c.)}\end{minipage} 
    \hspace{0.3cm}
    \includegraphics[width=.19\columnwidth]{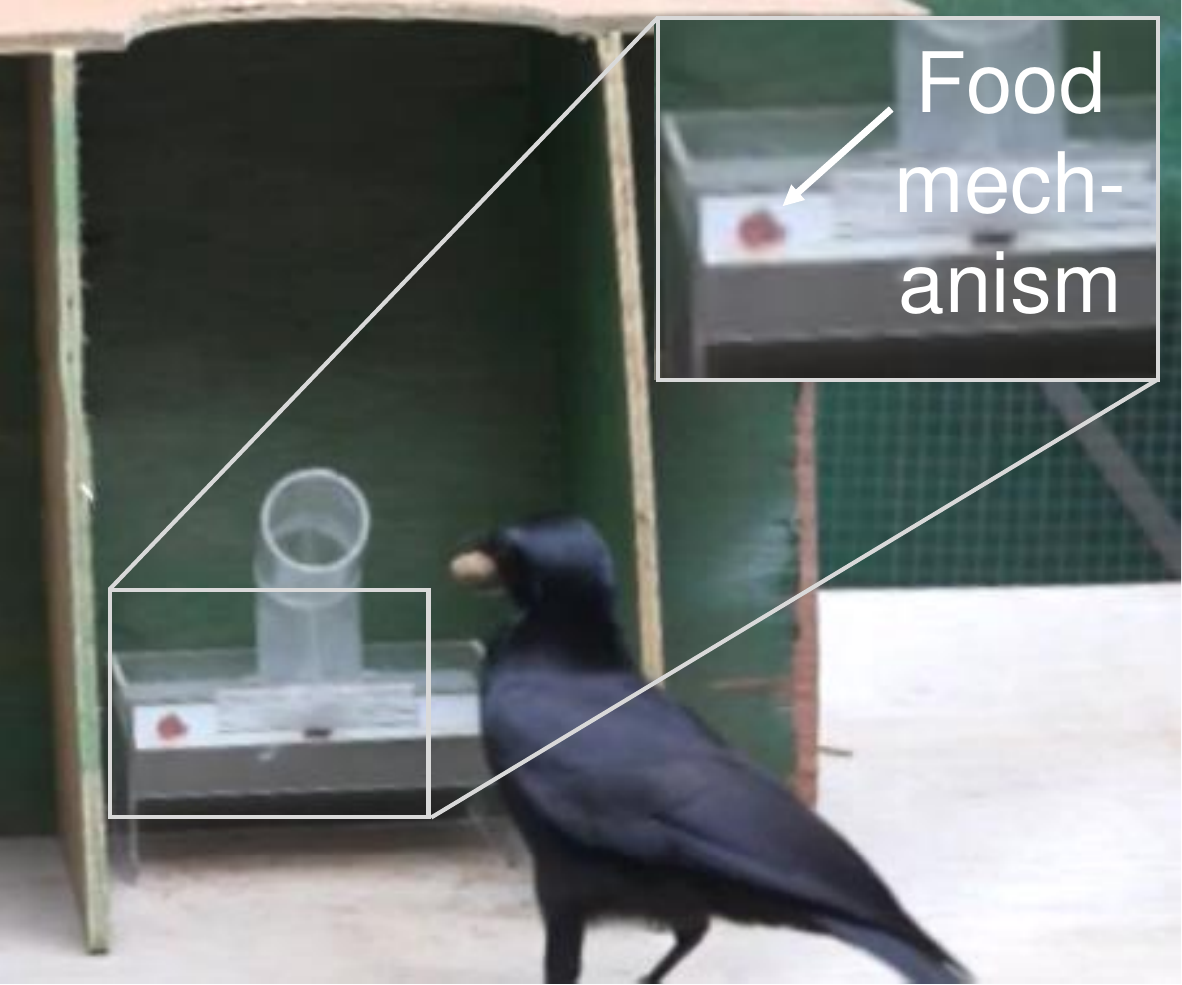}
    \hspace{-0.7cm}\begin{minipage}[b][0.3cm][t]{0.3cm}\textbf{(d.)}\end{minipage} 
    \hspace{0.3cm}
    \includegraphics[width=.19\columnwidth]{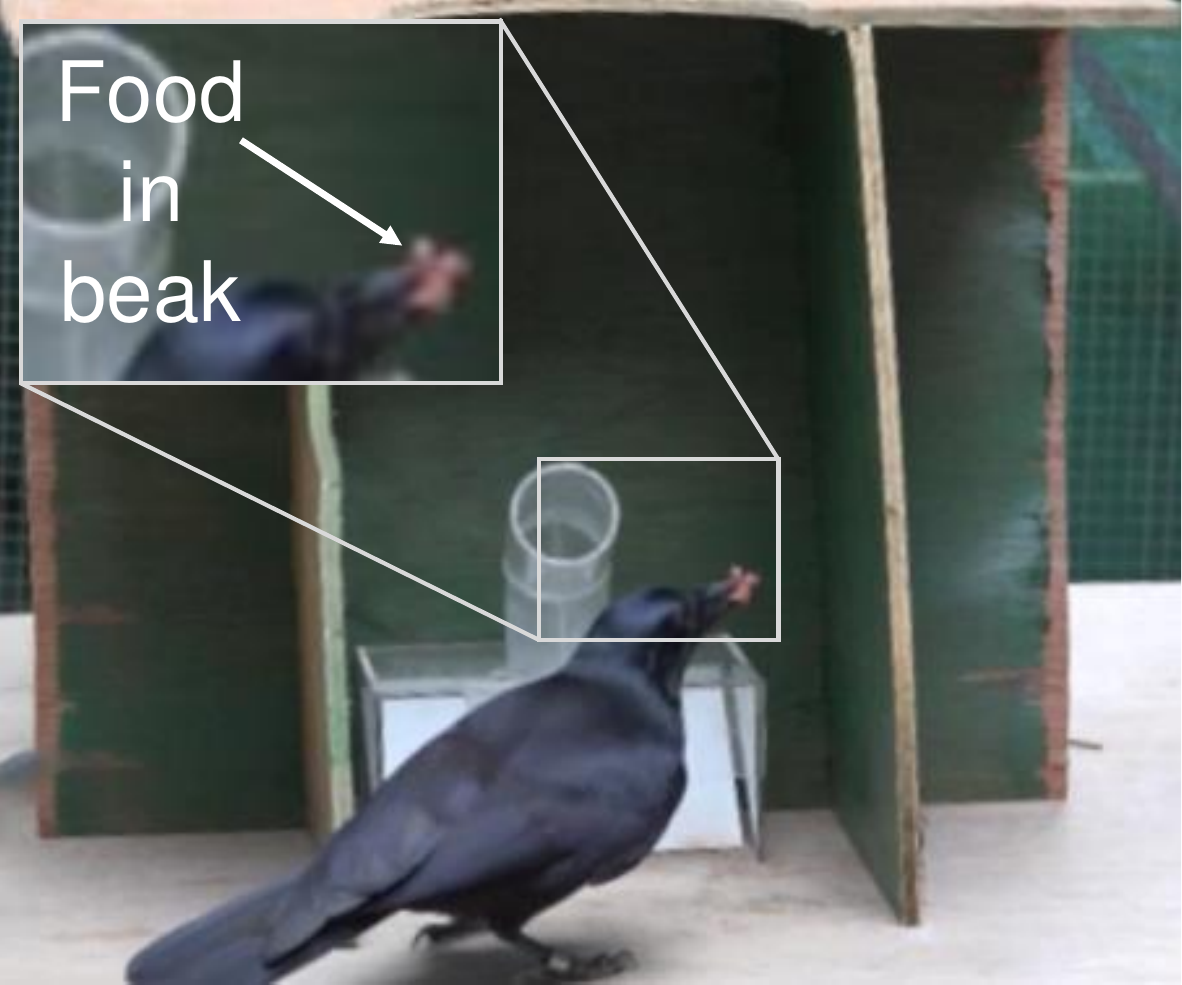}
    \hspace{-0.7cm}\begin{minipage}[b][0.3cm][t]{0.3cm}\textbf{(e.)}\end{minipage} 
    \hspace{0.3cm}
    \vspace{-10pt}

    \caption{
    \textbf{A New Caledonian crow solves a food-access problem.}
    \textbf{a}, First, the crow picks up a stick. \textbf{b-c}, Then it uses the stick to pull a stone out of a tube. \textbf{d-e,} Finally, it uses the stone to activate a mechanism that releases food. 
    Although the crow has never solved this problem setup before, it is able to solve it instantly after a brief inspection phase.
    Such zero-shot problem-solving is a prime example for \emph{hierarchical, goal-directed planning} that involves mental simulations and relies on a task-suitable, \emph{compositional abstractions}.
    The authors reason that the crow \emph{plans ahead} because it can only see one side of the cubic setup at a time.\cite{Gruber2019_CrowMentalRep} 
    Hence, the crow must \emph{mentally simulate} the future steps. 
    There is also \emph{transfer learning} involved: previously learned solutions to similar sub-problems are transferred and combined to solve the new problem. 
    Successful planning and transfer learning can only be expected to occur once \emph{abstract compositional mental representations} have developed, which, for example, may encode tool manipulation options. 
    To learn different tool manipulations, however, the crow relies on its \emph{intrinsic motivation} to acquire knowledge by playing and experimenting with the tools. 
    The images are used with permission by R. Gruber and co-authors\cite{Gruber2019_CrowMentalRep}.
    }.
    \label{fig:crow_probsolving}
\end{figure*}

Humans and several other intelligent animal species have the ability to break down complex problems into simpler, previously learned sub-problems. This hierarchical approach allows them to solve previously unseen problems in a zero-shot manner, i.e., without any trial and error. 
For example, \autoref{fig:crow_probsolving} depicts how a crow solves a non-trivial food-access puzzle that consists of three causal steps: It first picks a stick, then uses the stick to access a stone, and then uses the stone to activate a mechanism that releases food \cite{Gruber2019_CrowMentalRep}. 
There exist numerous analogous experiments that attest similar capabilities to primates, octopuses, and, of course,  humans \cite{Butz2017_MindBeing,Perkins1992}.
Human and animal cognition research suggests that hierarchical learning and problem-solving is critical to develop such skills \cite{Botvinick2009_HierarchicalCognitionRL,Butz2017_MindBeing,Tomov2020}. 
This raises the question of how we can equip intelligent artificial agents and robots with similar hierarchical learning and zero-shot problem-solving abilities.

We address this question from the perspective of reinforcement learning (RL) \cite{Arulkumaran2017_DeepRL,Li2018_DeepRL_Overview,Sutton:2018}. 
Several studies suggest that RL is biologically and cognitively plausible \cite{Neftci2019_RL_BioArtificial_NatureMI,Sutton:2018}. Many existing RL-based methods are aiming at zero-shot problem-solving and transfer learning, but this is currently only possible for minor variations of the same or a similar task \cite{Eppe2019_planning_rl} or in simple synthetic domains, such as 2D grid worlds \cite{Oh2017_Zero-shotTaskGeneralizationDeepRL,Sohn2018_Zero-ShotHRL}.
A continuous-space problem-solving behaviour that is comparable with the crow's behaviour of \autoref{fig:crow_probsolving} has not yet been accomplished with any artificial system.
We suggest that the focus on hierarchical learning and problem-solving is largely missing, seeing 
that research in hierarchical reinforcement learning (HRL) is underrepresented, as suggested by our meta review in \autoref{box:appendix:meta_RL-survey} of the supplementary material.

In this article, we address this gap in three steps: First, we evaluate the neurocognitive foundations of hierarchical decision-making and identify important mechanisms that enable advanced problem-solving skills. 
Second, we reveal an integration problem of contemporary hierarchical RL methods: We show that the biological mechanisms have mostly been implemented already---but only in isolation and not in an integrative manner. 
Third, we suggest steps towards combining key methods and mechanisms to overcome the integration challenge for the development of a unifying cognitive architecture.


\begin{figure*}[ht!]
    \centering
        \includegraphics[width=\textwidth,trim={0pt 1cm 0pt 0pt},clip]{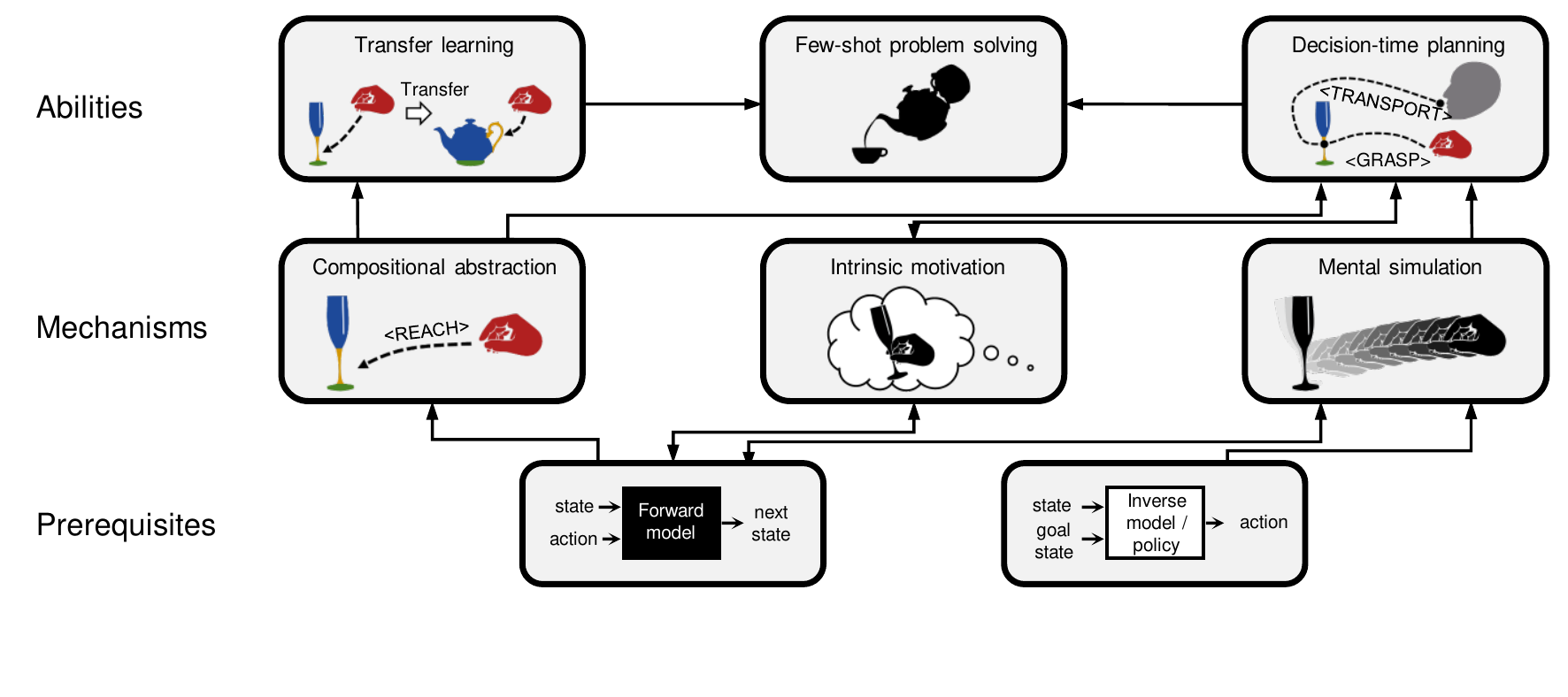}
    \caption{
    \textbf{Prerequisites, mechanisms, and features of biological problem-solving agents.}
    \emph{Forward-} and \emph{inverse models} are prerequisites for higher-order mechanisms and abilities.
    \emph{Compositional abstractions} allow the agent to decompose a problem into reusable substructures, e.g., by learning which part of an object is easily graspable, such as the stem of a glass.
    \emph{Intrinsic motivation} applied with the help of forward models can drive an agent to interact with its environment in an information-seeking, epistemic way, e.g., by eliciting interactions with the glass to learn more about its properties.
    \emph{Mental simulations} enable the exploration of hypothetical state-action sequences, e.g., by imagining how the hand position will evolve while reaching for the glass.
    Given these mechanisms, an agent can flexibly achieve desired goal states via hierarchical, compositional \emph{goal-directed planning}, e.g., planning to drink from the glass by first suitably grasping and then appropriately transporting it.
    Compositional abstractions enable the agent to perform \emph{transfer learning}, e.g., by executing suitable grasps onto other objects with stem-like parts, such as the shown teapot. 
    In summary, these abilities enable \emph{few-shot problem-solving}.}
    \label{fig:cog_prereq}
\end{figure*}

\section{Neurocognitive foundations}
\label{sec:cognitive_background}
In the following, we propose that the complex problem-solving skills in biological agents can be distinctively characterised by particular cognitive abilities. 
In our perspective, these abilities depend on key cognitive mechanisms, including  abstraction, intrinsic motivation, and mental simulation.
\autoref{fig:cog_prereq} outlines how these mechanisms may develop from forward and inverse models, as the fundamental neuro-functional prerequisites.

\subsection{Cognitive abilities}
As distinct cognitive abilities, we focus on the following traits and properties.

\smallskip
\noindent
\textbf{Few-shot problem-solving} is the ability to solve unknown problems with few ($\lessapprox 5$) trials. 
Zero-shot problem-solving is a special case of few-shot problem-solving, where no additional training at all is required to solve a new problem.
%
For example, a crow can use a stick as a tool for a novel food-access problem without further training (see \autoref{fig:crow_probsolving}), given it has previously solved related problems \cite{Gruber2019_CrowMentalRep}.
We identify two abilities that are most central for such problem-solving, namely transfer learning and decision-time planning (see \autoref{fig:cog_prereq}). 

\smallskip
\noindent
\textbf{Transfer learning}
 allows biological agents to perform few-shot pro\-blem-solving by transferring the solution of a previously solved task to novel, but analogous, unseen tasks \cite{Perkins1992}. This significantly reduces and sometimes eliminates the number of trials to solve a problem.

Such analogical reasoning is often considered as a critical cornerstone of human cognition.
For example, cognitive theory suggests that humans  understand mechanical systems by using analogous knowledge from other domains\cite{Hegarty2004_MechReasMentalSim}.
And education theory suggests that human transfer learning improves when explicitly trained to identify analogies between different problems\cite{Klauer1989_AnalogyTransfer}.
However, while humans seem skilful in applying analogies for problem-solving, it seems to be more difficult to discover new ones. 
For example, the phenomenon of functional fixedness describes the human problem-solving tendency to use objects only in the way they are traditionally used \cite{Duncker1945}.

\smallskip
\noindent
\textbf{Goal-directed planning.}
Behaviour is traditionally divided into two categories: Stimulus-driven, habitual behaviour and goal-directed, planned behaviour\cite{Dayan2009, Dolan2013, ODoherty:2017}.
Habitual behaviour refers to forms of reflexive control that are strongly automated, computationally efficient, and mainly learned from past reinforcements \cite{Dolan2013}.
One of the first critics of this approach was Tolman, who showed that rats can find rewards in a maze faster when they have visited the maze before, even if their previous visit had not been rewarded \cite{Tolman1930,Butz:2002c}.
The results suggest that the rats form a representation, or cognitive map, of the maze, which enables them to plan their behaviour once a reward was detected.
Advances in AI and cognitive science have influenced one another to understand goal-directed planning.
An influential model for action control was the TOTE model \cite{Miller:1960}, where actions are selected based on an iterative, hierarchical feedback loop. Herein, a desired anticipated state is compared to the current sensory state to select the next action.
Similar to TOTE, many recent theories view planning as a hierarchical process with different levels of abstraction \cite{Botvinick2014, Tomov2020, Wiener2003}.

But what kind of information is considered while planning?
Ideomotor theories of action generation \cite{Stock2004_Ideomotor} suggest that anticipated action effects determine action selection.
Accordingly, the theory of event coding \cite{Hommel2001} claims that representations of actions and their effects are stored in a common code, whereby action effect anticipations can directly activate corresponding actions \cite{Hommel2001}.
Various experimental evidence supports this link between action generation and effect prediction \cite{Hommel2001, Hoffmann:2003, Kunde:2007}.
Our supplementary \autoref{box:appendix:automatization} describes how the continuous learning of action-effect associations questions the traditional dichotomy between planned and habitual behaviour\cite{Dolan2013}.

\subsection{Cognitive mechanisms for transfer learning and planning}
The previous paragraphs discussed how transfer learning and planning enable intelligent few-shot problem-solving. But what enables transfer learning and planning? 
Our survey suggests three critical mechanisms for transfer learning and planning, namely sensorimotor abstraction, intrinsic motivation, and mental simulation.

\begin{figure}
\begin{minipage}{.32\textwidth}
\centering
\includegraphics[width=\textwidth]{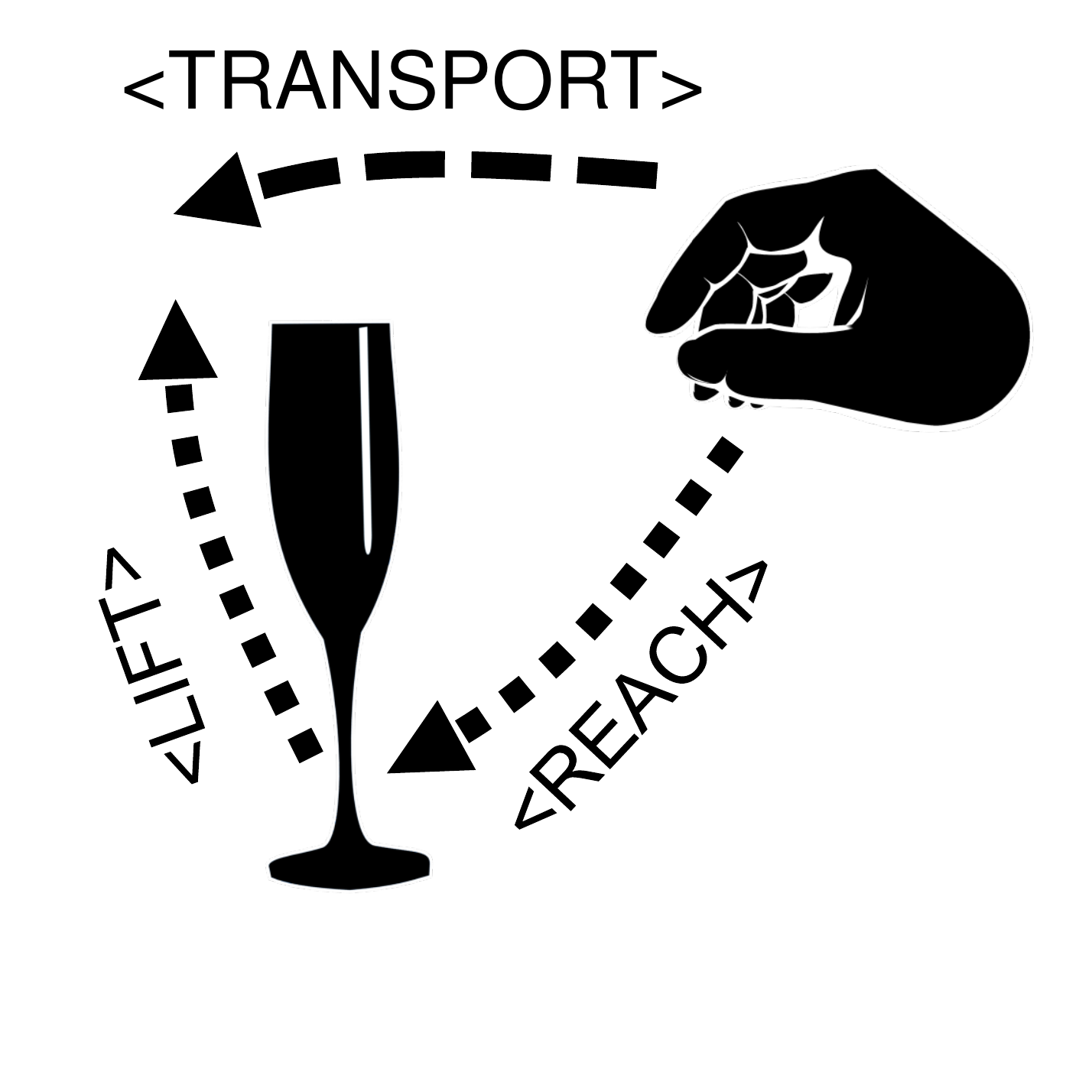}
\hspace{-0.8cm}\begin{minipage}[b][0.3cm][t]{0.3cm}(a.)\end{minipage} 
\end{minipage}
\begin{minipage}{.32\textwidth}
\centering
\includegraphics[width=\textwidth]{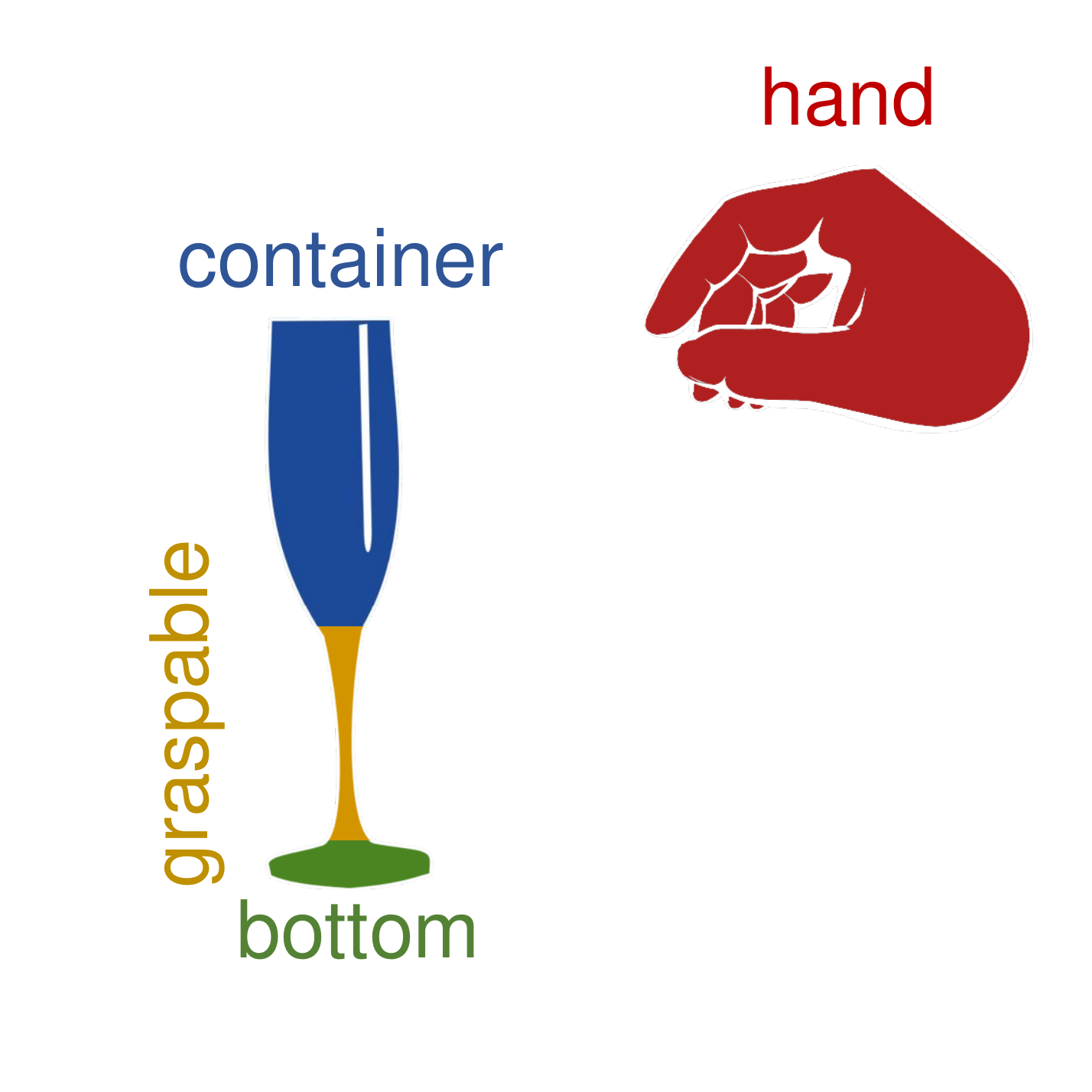}
\hspace{-0.8cm}\begin{minipage}[b][0.3cm][t]{0.3cm}(b.)\end{minipage} 
\end{minipage}
\hfill
\begin{minipage}{.32\textwidth}
\centering
\includegraphics[width=\textwidth]{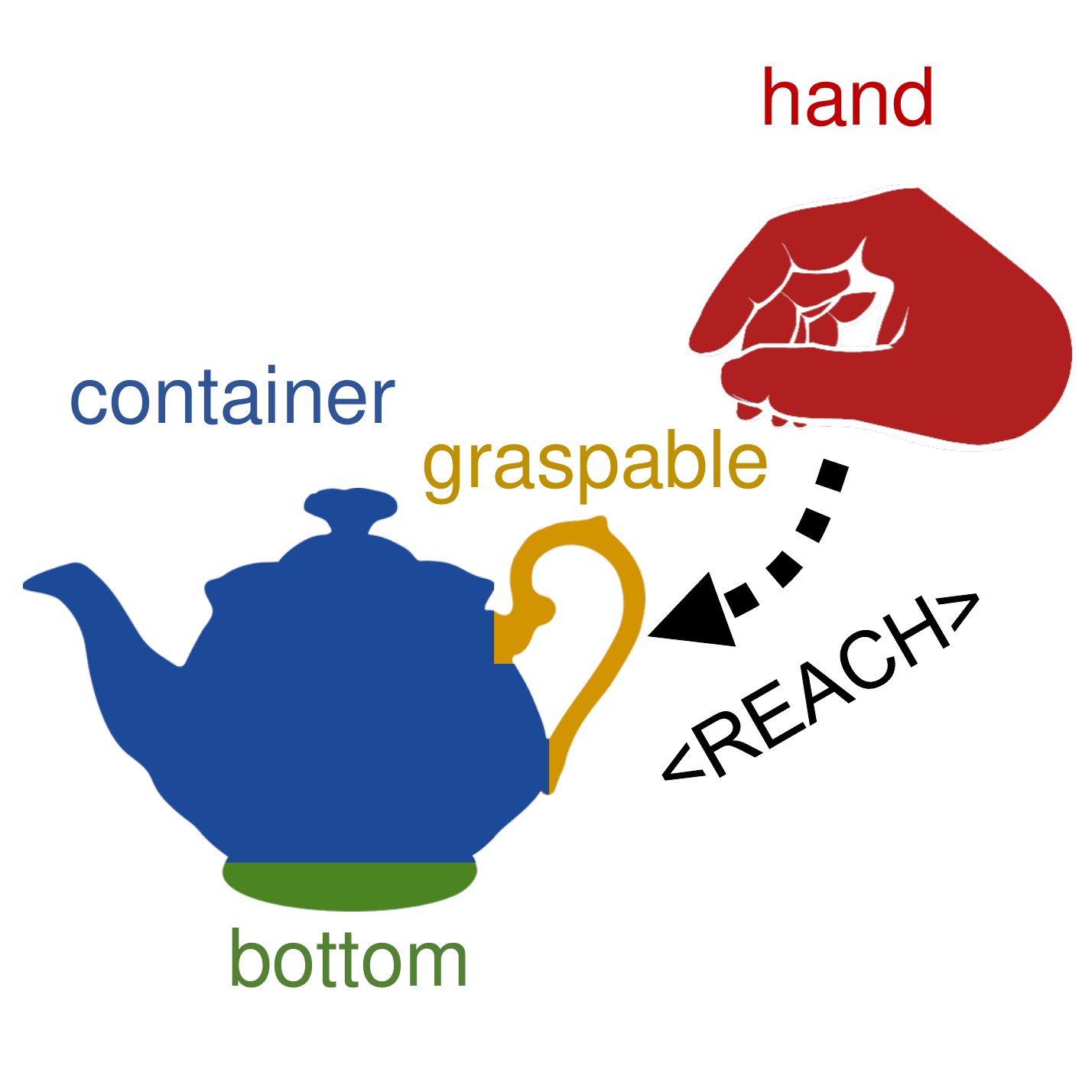}
\hspace{-0.8cm}\begin{minipage}[b][0.3cm][t]{0.3cm}(c.)\end{minipage} 
\end{minipage}
\caption{
\textbf{Compositional action and state abstractions.}
\textbf{a},  \emph{Action abstractions} describe a sequence of more primitive actions. \textbf{b}, \emph{State abstractions} encode certain properties of the state space. \textbf{c}, Their \emph{compositionality} enables their general application, by instantiating abstract definitions (the graspable target of reaching), with a specific object (teapot).}
\label{fig:cog_comp}
\end{figure}

\textbf{Sensorimotor abstraction.}
According to theories of embodied cognition, abstract mental concepts are inferred from our bodily-grounded sensorimotor environmental interaction experiences \cite{Barsalou2008_GroundedCognition,Butz2017_MindBeing,Butz2016,Pulvermuller:2010}. 
Herein, we differentiate between \emph{action abstraction} and \emph{state abstraction}.
Action abstractions refer to a temporally extended sequence of primitive actions, sometimes also referred to as options\cite{Sutton1999_options} or movement primitives\cite{Flash2005_AnimalMotorPrimitives, Schaal_2006}.
For example, \emph{transporting an object} is as an action abstraction since it is composed of a sequence of more primitive actions, such as \textit{reaching} and \textit{lifting} (see \autoref{fig:cog_comp}). 

State abstractions refer to reduced encodings of (parts of) the environment, where irrelevant details are abstracted away.
Accordingly, state abstractions may decompose a scene into task-critical aspects, such as graspability, containment, or hollowness.
For example, both \textit{wine glasses} and \textit{tea pots} have the property of being a \textit{container} and being \textit{graspable} (see \autoref{fig:cog_comp}).
Meanwhile, a \textit{container} implies \textit{hollowness}, etc.---illustrating the hierarchical and \textbf{compositional} nature of such abstractions. 
Cognitive linguistics and related fields suggest that abstractions benefit from compositionality \cite{Feldman2009_neural_ECG}. 
Formally, expressions are compositional if they are composed of sub-expressions and rules to determine the semantics of the composition. 
The language of thought theory\cite{Fodor:2001} transfers the compositionality principle from language to abstract mental representations, claiming that thought must also be compositional. 
For example, humans can easily imagine previously unseen images by composing their known constituents from linguistic descriptions \cite{Frankland2020_SearchLoT}, even fantastic ones: think of how a pink dragon riding a skateboard would look like. 

It is still being debated whether the abstract mental representations are encoded as local symbols or by means of distributed encodings \cite{Hummel2011_Symbols_Neural}, but in both cases there is evidence for the existence of combinatorial generalisation.
For example, \citet{Haynes2015_CompositionalNeuroRep} show that the neural codes in the ventrolateral prefrontal cortex, which encode compound behavioural rules, can be decomposed into neural codes that encode their constituents.

For sensorimotor abstractions, the rules of composition need to be flexibly inferred by our minds to form plausible environmental scenarios in accordance with our world knowledge \cite{Butz2016,Gaerdenfors:2014,Lakoff1999}.
Such \emph{common sense compositionality} makes abstractions applicable in meaningful manners. 
The filling in of an abstract component (e.g., the  target of a grasp) is constrained towards applicable objects (e.g., a graspable object such as a teapot), as depicted in \autoref{fig:cog_comp}.
From an algorithmic perspective, common sense compositionality brings enormous computational benefits:
Abstractions, e.g., a grasping-encoding, are more economical if they can be applied in various situations, e.g. for different grasping targets.
Common sense compositionality essentially simplifies the identification of analogies by enabling mappings between representations, an advantage well-known in cognitive theories of creativity and concept blending\cite{Eppe2018_AIJ_Blending,Turner2014} (cf. supplementary \autoref{box:appendix:compositionality-example}).

\smallskip
\noindent
\textbf{Intrinsic motivation} affects goal-directed reasoning by setting intrinsically motivated goals.  
From a cognitive development perspective, the term \emph{intrinsic motivation} was coined to describe the ``inherent tendency to seek out novelty, [...] to explore and to learn''\cite{RyanDeci2000}.
Here, \emph{intrinsic} is used in opposition to extrinsic values, which are rather associated with the satisfaction of needs or the avoidance of undesired outcomes \cite{Friston2015}.
Intrinsic motivations induce a general exploratory behaviour, curiosity, and playfulness.
Curiosity refers to information gain-oriented exploratory behaviour\cite{Berlyne1966, Loewenstein1994, Oudeyer2007}.
Curiosity is closely related to playfulness, where novel interactions are tried-out in a skill-improvement-oriented manner. Playfulness can be observed in certain intelligent animal species, such as dogs and corvids\cite{Pisula2008}.

\smallskip

\noindent
\textbf{Mental simulation} describes the ability of an agent to anticipate the dynamics of its environment on various representational levels. 
For example, \emph{motor-imagery} is deemed to enable athletes to improve the execution of challenging body movements (e.g. a backflip) by mentally simulating the movement\cite{Jeannerod1995_MentalImageryMotor}, and engineers can rely on mental simulation when developing mechanical systems \cite{Hegarty2004_MechReasMentalSim}. 
Several researchers suggest that mental simulation also takes place on higher reasoning levels. For example, \citet{Kahnemann1982_MentalSim}, \citet{Wells1989_MentalSimCausality}, and later \citet{Taylor1998_MentalSim} report how mental simulation improves planning of future behaviour on a causal conceptual level. 

\smallskip

\subsection{Forward-inverse models as functional prerequisites}

Sensorimotor abstraction, intrinsic motivation, and mental simulation are complex mechanisms that demand a suitable neuro-functional basis. 
We propose that this essential basis is constituted by forward and inverse models \cite{Stock2004_Ideomotor}. 
Forward models predict how an action affects the world, while inverse models determine the actions that need to be executed to achieve a goal.

\smallskip
\noindent
\textbf{Forward-inverse models for mental simulation.}
To perform mental simulation, an agent needs a forward model to simulate how the dynamics of the environment behave, possibly conditioned on the agent's actions.
However, the prediction of all consequences of all possible actions quickly becomes computationally intractable. 
Hence, action-selection requires more direct mappings from state and desired inner mental state to potential actions. 
This can be accomplished by an inverse model or behaviour policies, which generate candidate actions for accomplishing a current goal.

\smallskip

\noindent
\textbf{Forward-inverse models for intrinsic motivation.}
Forward and inverse models are also needed for intrinsic motivation. 
\rev{For example, the learning progress of internal forward models has been demonstrated to be useful for modeling curiosity as an intrinsic reward signal\cite{Kaplan:2004, Schmidhuber2010, Oudeyer2007}.}{R2-\ref{rev:r2:clarify_pred_errors}}
Friston et al.~propose that intrinsically motivated behaviour emerges when using forward models to apply \emph{active inference}\cite{Friston2011, Friston2015}, i.e., the process of inferring actions to minimise expected free energy.
The authors argue that active inference can lead to epistemic exploration, when an agent acts out different behaviour to reduce the uncertainty about their forward predictions \cite{Friston2015, Oudeyer2018_CompuCuriosityLearning}.
However, active inference and the free energy principle has been criticised as overly general, and its falsifiability has been questioned\cite{Colombo:2018}.

\smallskip
\noindent
\textbf{Forward-inverse models for abstraction and grounding.}
Various cognitive theories, including predictive coding\cite{Huang2011}, the free energy principle\cite{Friston2010}, and the Bayesian brain hypothesis\cite{Knill2004}, have viewed the brain as a generative machine, which constantly attempts to match incoming sensory signals with its probabilistic predictions\cite{Clark2013}.
Within these frameworks, prediction takes place on multiple processing hierarchies that interact bidirectionally:
Top-down information provides additional information to predict sensory signals, while bottom-up error information is used to correct high-level predictions \cite{Clark2016_SurfUncertainty}.
\citet{Butz2016} proposes that this hierarchical processing leads to the emergence of abstract, compositional encodings on higher processing levels, e.g., a container-concept, which enable predictions of lower-level features, e.g., the occlusion and transport of a contained object.
Event segmentation theory (EST)\cite{Zacks2007_EST}  makes the role of forward predictions for learning sensorimotor abstractions even more explicit:
According to EST, transient prediction errors are used to segment the continuous stream of experience into conceptual event-encodings, as detailed in the supplementary \autoref{box:appendix:EST}.


\section{Computational realisations}
\label{sec:hrlFinal}

\begin{figure}
    \centering
    \includegraphics[width=\textwidth]{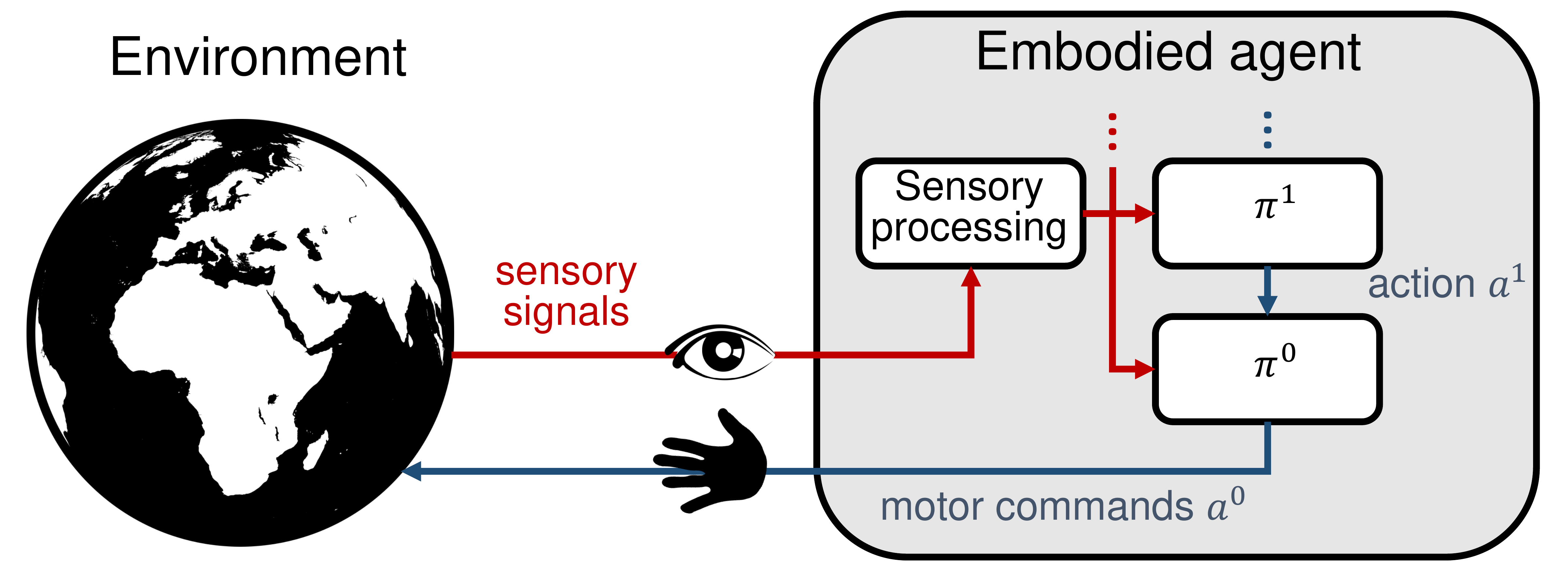} 
    \caption{
    \textbf{General hierarchical problem-solving architecture.}
    A separate policy $\pi^i$ is used for each layer of abstraction. 
    Most existing computational HRL approaches focus on two layers, where the high-level action $a^1$ is either a behavioural primitive or a subgoal. 
    The low-level actions $a^0$ are motor commands that affect the environment directly. }
    \label{fig:hrl}
\end{figure}

Computational hierarchical reinforcement learning (HRL) mechanisms (cf.~\autoref{fig:hrl}) are less integrated than those of biological agents. 
In the following, we summarize the state of the art of computational approaches in relation to the neurocognitive foundations. We provide an additional overview in the supplementary  \autoref{tab:hrl_properties}.

\subsection{Transfer learning and planning for few-shot abilities}
\label{sec:hrl:abilities}
Our survey of the neurocognitive foundations indicates that two foundational cognitive abilities for few-shot problem-solving are transfer learning and  planning. But how can we model these abilities computationally? 

\smallskip
\noindent
\textbf{Transfer learning} is natural for hierarchical architectures in the sense that they can potentially re-use general high-level skills and apply them to different low-level use cases, and vice versa. 
Therefore, a significant fraction of the existing transfer learning approaches build on learning re-usable low-level policies \cite{Eysenbach2019_DiversityFunction,Frans2018_MetaLearningHierarchies,Heess2016_transferModulControl,Jiang2019_HRL_Language_Abstraction,Li2020_SubPolicy,Qureshi2020_CompAgnosticPol,Sharma2020_DADS,Tessler2017_LifelongHRLMinecraft,Vezhnevets2016_STRAW}.
Research has not only considered transfer learning between different tasks, but also between different robot and agent morphologies\cite{Devin2017_Transfer_RL,Frans2018_MetaLearningHierarchies,Hejna2020_MorphologicalTransfer}. 

\smallskip
\noindent
\textbf{Planning} can be categorised into \emph{decision-time planning} and \emph{background planning}\cite{Hamrick2021_MBRL,Sutton:2018}. 
Decision-time planning refers to a search over sequences of actions to decide which action to perform next to achieve a specific goal.
The search is realised by simulating the actions with an internal predictive model of the domain dynamics.  
Decision-time planning enables zero-shot problem-solving because the agent only executes the planned actions if an internal prediction model verifies that the actions are successful.
Background planning refers to model-based reinforcement learning as introduced by \citet{Sutton1990_Dyna}. It is used to train a policy by simulating actions with a forward model.
This can improve the sampling efficiency, but not to the extent that it enables few-shot problem-solving.

Hierarchical decision-time planning is a well-known paradigm in classical artificial intelligence\cite{Nau2003_SHOP2}, but approaches that integrate it with hierarchical reinforcement learning are rare. 
Some approaches integrate action planning with reinforcement learning by using an action planner for high-level decision-making and a reinforcement learner for low-level motor control\cite{Eppe2019_planning_rl,Lyu2019_SDRL,Ma2020_MultiAgentHRL,Sharma2020_DADS}. 
These approaches underpin that decision-time planning is especially useful for high-level inference in discrete state-action spaces. 


\subsection{Mechanisms behind transfer learning and planning}
\label{sec:hrl:mechanisms}
Our summary of the cognitive principles behind transfer learning and planning reveals three important mechanisms that are critical for the learning and problem-solving abilities of biological agents, namely compositional sensorimotor abstraction, intrinsic motivation, and mental simulation.

\begin{figure}
    \centering
    \includegraphics[width=.99\linewidth,trim={1.5cm 0cm 1.5cm 0.0cm},clip]{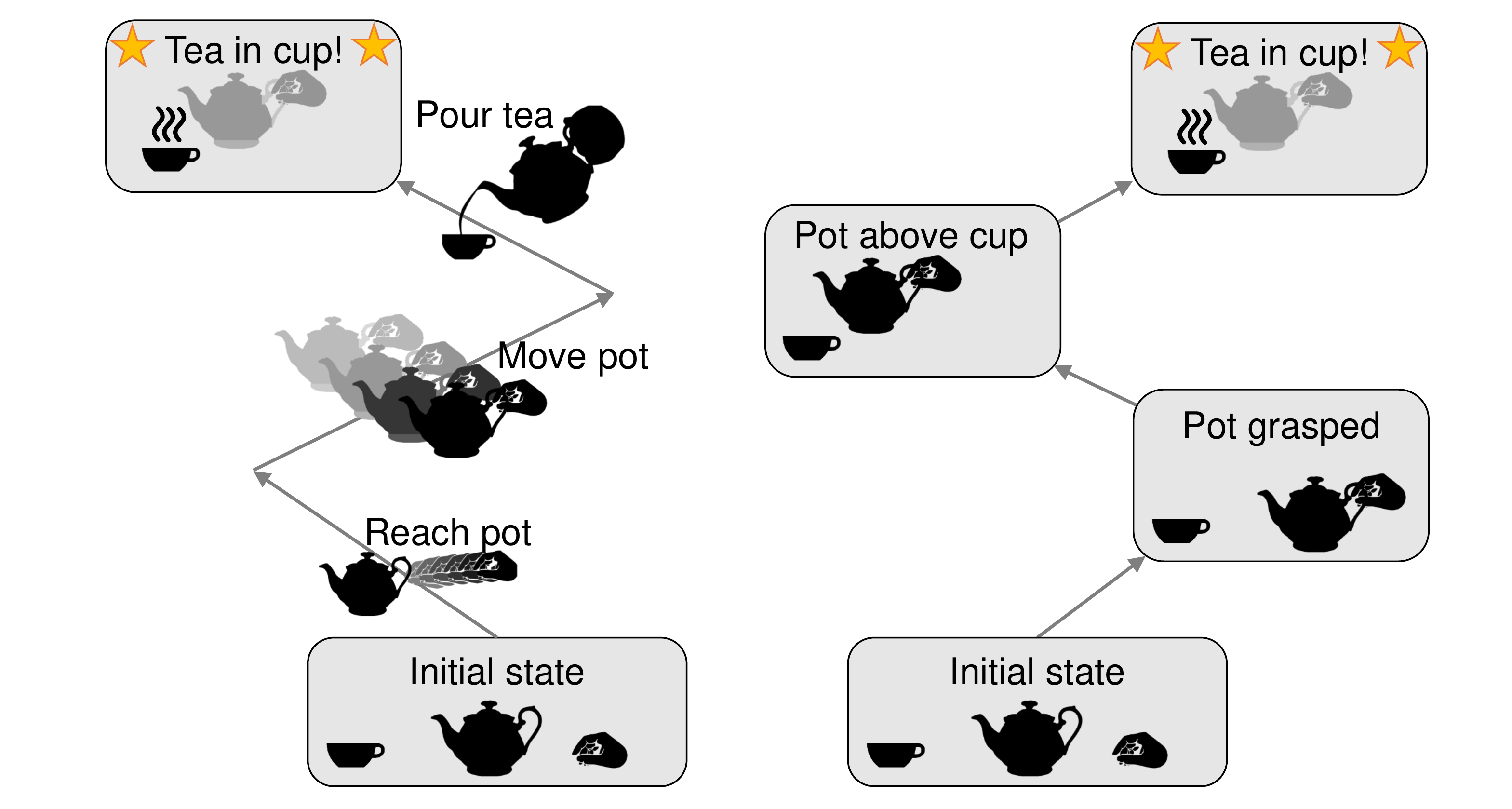}
    {\sf \footnotesize ~~~~~~~ (a.)    Behaviour primitives  ~~~~~~~~~~~~ (b.) Subgoals ~~~~~~~~~~~~ }
    \caption{
    \textbf{Types of action abstractions.}
    \textbf{a}, With behaviour primitives, the agent determines the path to the final goal by selecting a sequence of high-level actions, but without specifying explicitly to which intermediate state each action leads. 
    \textbf{b}, With subgoals, the agent determines the intermediate states as subgoals that lead to the final goal, but without specifying the actions to achieve the subgoals.}
    \label{fig:options_behaviour_primitives}
\end{figure}

\smallskip
\noindent
\textbf{Compositional sensorimotor abstraction and grounding.}
Hierarchical reinforcement learning (HRL) has a temporal and a representational dimension, and one can distinguish between action and state abstraction. 
The temporal abstraction of actions breaks down complex problems into a hierarchy of simpler problems. 
The probably most influential method for representational action abstraction builds on behaviour primitives
\cite{Bacon2017_OptionCritic,Dietterich2000_StateAbstraction_MAXQ,Frans2018_MetaLearningHierarchies,Kulkarni2016_HDQN,Li2020_SubPolicy,Shankar2020_MotorPrimitives,Sharma2020_DADS,Vezhnevets2020_OPRE}
, including options\cite{Sutton1999_options}, sub-policies, or atomic high-level skills.
Such behaviour primitives are represented in an abstract representational space, e.g. in a discrete finite space of action labels or indices (see \autoref{fig:options_behaviour_primitives}, a.). 
A more recent type of high-level action representations are subgoals in the low-level state-space \cite{Eppe2019_planning_rl,Ghazanfari2020_AssociationRules,Hejna2020_MorphologicalTransfer,Levy2019_Hierarchical,Nachum2018_HIRO,Rafati2019_model-free_rep_learning,Roeder2020_CHAC,Zhang2020_AdjancentSubgoals} (see \autoref{fig:options_behaviour_primitives}, b.).
Here, the agent achieves the final goal by following a sequence of subgoals.
Our supplementary \autoref{tab:hrl_properties} shows that  there is considerably less contemporary research on state abstraction than on action abstraction. 

Many current reinforcement learning approaches based on visual input use convolutional neural networks to perform a rather generic form of  abstraction, where all hierarchical layers use the same abstract visual feature map\cite{Jiang2019_HRL_Language_Abstraction,Lample2018_PlayingFPSGames,Oh2017_Zero-shotTaskGeneralizationDeepRL,Sohn2018_Zero-ShotHRL,Vezhnevets2017_Feudal,Wulfmeier2020_HierachicalCompositionalPolicies,Yang2018_HierarchicalControl}. 
A problem is that this does not appreciate that \emph{different levels of inference require different levels of abstraction}: For example,  for a transport task, the exact shape and weight of an object is only relevant for the low-level motor control. A high-level planning layer only requires abstract facts, such as whether the object is graspable or transportable.
Therefore, the representational state abstraction should fit the level of inference it is used for. We refer to such abstraction as \textbf{inference-fitted abstraction}. 
%
Inference-fitted abstraction has been tackled in existing HRL approaches, but many of them rely on manually defined abstraction functions\cite{Eppe2019_planning_rl,Kulkarni2016_HDQN,Ma2020_MultiAgentHRL,Toussaint2018}.
There exist only a few exceptions where state abstractions are derived automatically in hierarchical reinforcement learning%
, e.g. through  clustering\cite{Akrour2018_RL_StateAbstraction,Ghazanfari2020_AssociationRules}, with feature representations\cite{Schaul2013_Better_Generalization_with_Forecasts}, or by factorisation \cite{Vezhnevets2020_OPRE}.

Cognitive science also suggests that \emph{compositionality} is an important property of abstract representations. 
For instance, a symbolic compositional action representation \concept{grasp(glass)} allows for modulating the action \emph{grasp} with the object \emph{glass}.
Compositionality is also applicable to distributed numerical expressions, such as vector representations. For example, a vector $v_1$ is a compositional representation if it is composed of other vectors, e.g., $v_2$ and $v_3$, and if there is a vector operation $\circ$ that implements interpretation rules, e.g., $v_1 = v_2 \circ v_3$. 

There is significant cognitive evidence that compositionality improves transfer learning\cite{Colas2020_LanguageGroundingRL}. 
This evidence is computationally plausible when considering that transfer learning relies on exploiting analogies between problems, as described in the supplementary \autoref{box:appendix:compositionality-example}
Other cognitively inspired computational methods, such as concept blending\cite{Eppe2018_AIJ_Blending}, provide further evidence that a lower number of possible analogy mappings improves transferability.
Furthermore, the work by \citet{Jiang2019_HRL_Language_Abstraction} provides empirical evidence that compositional representations improve learning transfer. The authors use natural language as an inherently compositional representation to describe high-level actions in hierarchical reinforcement learning, and show that this natural language representation improves transfer learning performance compared to non-compositional representations.

\smallskip
\noindent
\textbf{Intrinsic motivation}
is a useful method to stabilise reinforcement learning by supplementing sparse extrinsic rewards with additional  domain-independent intrinsic rewards. 
The most common method of HRL with intrinsic motivation is to provide intrinsic rewards when subgoals or subtasks are achieved%
\cite{Blaes2019_CWYC,Eppe2019_planning_rl,Haarnoja2018_LatentSpacePoliciesHRL,Kulkarni2016_HDQN,Lyu2019_SDRL,Oh2017_Zero-shotTaskGeneralizationDeepRL,Qureshi2020_CompAgnosticPol,Rasmussen2017_NeuralHRL,Riedmiller2018_SAC-X,Yang2018_PEORL}.

Other approaches provide intrinsic motivation to identify a collection of behavioural primitives with a high diversity\cite{Blaes2019_CWYC, Eysenbach2019_DiversityFunction,Machado2017_PVF_Laplace_OptionDiscovery} and predictability\cite{Sharma2020_DADS}. 
This includes also the identification of primitives suitable for re-composing high-level tasks\cite{Shankar2020_MotorPrimitives}.

Another prominent intrinsic reward model that is commonly used in non-hierarchical reinforcement learning is based on surprise and curiosity\cite{Oudeyer2007,Pathak2017_forward_model_intrinsic,Schillaci2020_IntrinsicMotivation,Schmidhuber2010}. 
In these approaches, surprise is often modelled as a function of the prediction error of a forward model, and curiosity is realised by providing intrinsic rewards if the agent is surprised.
However, only \citet{Blaes2019_CWYC}, \citet{Colas2019_CURIOUS}, and \citet{Roeder2020_CHAC} use surprise in a hierarchical setting, showing that hierarchical curiosity leads to a significant improvement of the learning performance.
Hierarchical curiosity provides potential synergies because a curious high-level layer automatically acts as a curriculum generator for a low-level layer. For example, it can generate ``Goldilocks'' subgoals that are neither too hard nor too easy to achieve for the low-level layer. 
The difficulty is to identify these Goldilocks conditions and to find an appropriate balance between explorative and exploitative subgoals.


\smallskip
\noindent
\textbf{Mental simulation} enables an agent to anticipate the effects of its own and other actions. Therefore, it is a core mechanism to equip an intelligent agent with the ability to plan ahead. Computational approaches for mental simulation involve model-based reinforcement learning\cite{Sutton1990_Dyna}, action planning\cite{Nau2003_SHOP2}, or a combination of both \cite{Eppe2019_planning_rl,Hafez2020_dual-system}. 
However, even though there is cognitive evidence that mental simulation happens on multiple representational layers\cite{Butz2017_MindBeing,Frankland2020_SearchLoT}, there is a lack of approaches that use hierarchical mental simulation. 
Only a few hierarchical approaches that integrate planning with reinforcement learning build on mental simulation\cite{Eppe2019_planning_rl,Lyu2019_SDRL,Ma2020_MultiAgentHRL,Sharma2020_DADS,Yamamoto2018_RL_Planning,Yang2018_PEORL}, though the mental simulation mechanisms of these models are only implemented on the high-level planning layer. 
An exception is the work by \citet{Wu2019_ModelPrimitives}, who use mental simulation on the low-level layer to determine the sub-policies to be executed. 
Another exception is presented by \citet{Li2017_efficient_learning}, who perform mental simulation for planning on multiple task layers. 
However, we are not aware of any approach that performs hierarchical background planning, i.e., that trains a hierarchical policy by mentally simulating the action execution on multiple levels.
One possible reason for this lack is that high-level background planning requires stable low-level representations, and it is difficult to model a respective stability criterion.



\subsection{Prerequisites for sensorimotor abstraction, intrinsic motivation, and mental simulation}
Reinforcement learning builds on policies or inverse models that predict the actions to be executed in the current state to maximise reward or to achieve a goal state.
In contrast, a forward model predicts a future world state based on the current state and a course of actions. 
Our review shows that the vast majority of hierarchical reinforcement learning methods use inverse models for both the high-level and low-level layers. 
Some approaches exist that use an inverse model only for the low-level layer\cite{Eppe2019_planning_rl,Lyu2019_SDRL,Ma2020_MultiAgentHRL,Sharma2020_DADS,Yamamoto2018_RL_Planning,Yang2018_PEORL} and use decision-time planning to perform high-level decision-making.
Our review also implies that a forward model is required for several additional mechanisms that are necessary or at least highly beneficial for transfer learning, planning, and few-shot problem-solving. 
Some non-hierarchical approaches use a forward model to perform sensorimotor abstraction\cite{Hafner2020_Dreamer,Pathak2017_forward_model_intrinsic}. They achieve this with a self-supervised process, where forward predictions are learned in a latent abstract space. 
However, we are not aware of any hierarchical method that exploits this mechanism. 

A forward model is also useful to model curiosity as an intrinsic motivation mechanism, for example, by rewarding an agent for being surprised, where surprise is a function of the prediction error \cite{Friston2011}. 
Curiosity has been demonstrated to alleviate the sparse rewards problem in reinforcement learning\cite{Pathak2017_forward_model_intrinsic,Roeder2020_CHAC}, but with few exceptions\cite{Colas2019_CURIOUS,Roeder2020_CHAC}, there is a lack of approaches that use a hierarchical forward model for generating hierarchical curiosity. 



\section{Critical discussion and outlook}
\label{sec:results}
\begin{figure*}
    \centering
    \includegraphics[width=\textwidth]{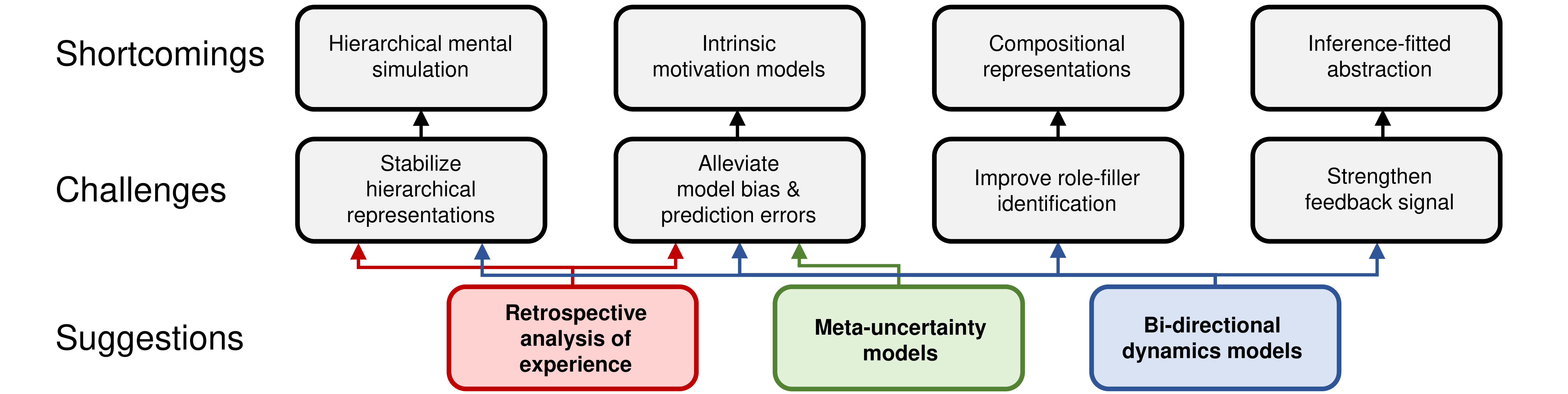}
    \caption{
    \textbf{Shortcomings, challenges, and suggestions for HRL.}
    Our section \nameref{sec:hrlFinal} identifies several shortcomings of contemporary HRL methods. We suppose that these result from at least four central challenges, and we provide three suggestions for key methods that may help to address the challenges.}
    \label{fig:challenges_shortcomings_solutions}
\end{figure*}

Based on our considerations above and the supplementary \autoref{box:appendix:hrl_review} and \autoref{tab:hrl_properties},
we now proceed with summarising the shortcomings of current hierarchical reinforcement learning approaches. 
The shortcomings are consequences of specific challenges that arise from the integration of the different mechanisms (see \autoref{fig:challenges_shortcomings_solutions}). Here, we address these shortcomings with specific algorithmic suggestions. 

\subsection{Shortcomings and challenges of HRL architectures}
\label{sec:results:challenges}
We identify the following major challenges for current methods:

\begin{sloppypar}
\noindent \textbf{Hierarchical mental simulation requires stable hierarchical representations and precise predictions.}
Most current few-shot problem-solving methods build on transfer learning, but they neither leverage decision-time planning nor background planning to exploit hierarchical mental simulation.
While numerous non-hierarchical methods show how mental simulation can improve sample efficiency\cite{Deisenroth2011,Ha2018_WorldModels}, our summary in the supplementary \autoref{tab:hrl_properties} shows that reinforcement learning with hierarchical mental simulation on multiple layers is implemented in only one recent method\cite{Li2017_efficient_learning}.
A major reason for this shortcoming is that the low-level  policy should have already converged before the high-level forward model starts training. 
This problem has been addressed for model-free HRL \cite{Eysenbach2019_DiversityFunction}, but it has not yet been investigated with  predictive forward models. 
\end{sloppypar}


\begin{sloppypar}
\noindent \textbf{Intrinsic motivation requires forward models with elaborate uncertainty-handling.}
Our supplementary \autoref{tab:hrl_properties} show that curiosity and diversity are underrepresented intrinsic motivation methods: they are implemented in only 9 out of 50 surveyed papers, while our section \nameref{sec:cognitive_background} shows significant cognitive evidence that these mechanisms are critical for intelligent behaviour in biological agents\cite{Oudeyer2018_CompuCuriosityLearning}.

A potential reason for this shortcoming is that prediction-based intrinsic motivation suffers from an overly simplistic uncertainty handling of current computational forward models.
This is an issue for noisy observations or chaotic and unpredictable system dynamics. 
For example, \citet{Roeder2020_CHAC} achieve curiosity-driven intrinsic motivation by maximising the prediction error of a forward model. 
This causes the agent to suffer from the so-called \emph{noisy TV problem}, where an agent always strives to encounter unpredictable observations (such as a TV showing only white noise). 
This problem interferes with the problem of learning hierarchical representations because it does not address uncertainties of high-level representations if the learning of these representations has not yet converged.
\end{sloppypar}

\noindent \textbf{Compositionality requires elaborate role-filler representations.}
Only a few HRL methods consider compositionality. 
Exceptions include natural language-based representations of actions \cite{Jiang2019_HRL_Language_Abstraction} or symbolic logic-based compositional representations \cite{Oh2017_Zero-shotTaskGeneralizationDeepRL}.
Since compositional representations generally build on variable-value representations, these need to be researched first on a more foundational level to alleviate this shortcoming. 
Graph neural networks, a class of neural networks that encode entities and their interactions as nodes and edges within a graph \cite{Battaglia:2018}, are a promising approach for compositional representations.
When provided with the underlying graph structure, Graph neural networks seem to generalise better and support relational reasoning \cite{Battaglia:2018}.
However, discovering a suitably abstracted graph from data remains an open challenge.

\noindent \textbf{Inference-fitted state abstraction requires a strong learning signal.}
Recent HRL approaches \cite{Levy2019_Hierarchical,Roeder2020_CHAC} use the same state representation for each layer in the hierarchy. 
However, it seems cognitively more plausible to fit the level of abstraction to the level of inference needed on each hierarchical level.
For example, a high-level layer may only need to know whether an object is graspable, it does not need to consider its precise shape. This more detailed information, however, is relevant for the low-level layer that performs the grasping.
There exist RL-methods that learn state abstractions from visual data with a convolutional autoencoder\cite{Ha2018_WorldModels}. 
However, such representations are optimised for reconstructing the visual input---not for decision-making. 
Furthermore, in the existing recent vision-based architectures\cite{Jiang2019_HRL_Language_Abstraction,Lample2018_PlayingFPSGames,Oh2017_Zero-shotTaskGeneralizationDeepRL,Sohn2018_Zero-ShotHRL,Vezhnevets2017_Feudal,Wulfmeier2020_HierachicalCompositionalPolicies,Yang2018_HierarchicalControl}, all hierarchical layers of the RL agent are provided with the same image embedding. 
Therefore, the abstractions are not inference-fitted. 
A problem for learning inference-fitted abstraction functions is that the reward signal of reinforcement learning is a relatively weak driver for the learning process, compared to other methods like supervised learning. 

\subsection{Outlook and suggestions for addressing the challenges}
\label{sec:results:solutions}
In the following, we sketch how computational key methods may potentially help to address the above challenges. 

\noindent \textbf{Retrospective analysis of experiences.}
A central problem of HRL is that stable high-level representations can only emerge once the learning of the lower-level has converged \cite{Eysenbach2019_DiversityFunction}. 
We suggest that a retrospective analysis of experiences can alleviate this issue.
This has been demonstrated for training a policy by learning in hindsight\cite{Andrychowicz2017,Levy2019_Hierarchical}, where the achieved world state of a policy is retrospectively pretended to be the desired goal state. 
We suggest that hindsight learning is a special case of a more general retrospective analysis of memorised experiences. 
For example, it is potentially possible to also learn hierarchical forward models in hindsight, without the high-level layer having to wait for a low-level layer to converge. 
We hypothesise that re-considering the outcome of past erroneous actions of a not yet fully trained low-level layer as a desired outcome can stabilise the high-level representations. 
Such a technique may also help to alleviate the model bias.

\noindent \textbf{Meta- uncertainty models.}
Prediction errors play an important role in mental simulation and intrinsic motivation. 
In our section \nameref{sec:cognitive_background} we have discussed the free energy principle, which involves two seemingly contradicting mechanisms that build on predictive processing models. 
On the one hand, there is active inference \cite{Schwartenbeck2019_CompUncertainExplo}, which seeks to minimise long-term surprise. This can be implemented by minimising a forward model's uncertainty or prediction error. On the other hand, there is active learning, which seeks to maximise information gain. This can be implemented by maximising uncertainty\cite{Haarnoja2018_SAC} or prediction errors \cite{Roeder2020_CHAC} through respective intrinsic reward functions.
\citet{Schwartenbeck2019_CompUncertainExplo} address this contradiction by building on \emph{expected uncertainty}\cite{Yu2005UncertaintyAttention} to unify both mechanisms. 
Expected as well as unexpected uncertainty then can be understood as a meta-uncertainty-analysis. 
We suppose that an intrinsic reward for high \emph{meta-}uncertainty may lead to exploratory behaviour, causing the exploration of high short-term but anticipated lower long-term meta-uncertainty. Furthermore, the gained certainty about the uncertainty of forward models can lead to improved predictive processes, which are aware of their own limits\cite{Baldwin:2021}. 

\noindent \textbf{Bidirectional dynamics models.}
Bidirectional models involve both forward and backward inference, and we suppose that this combination is beneficial for representational abstraction in
hierarchical architectures.
Our supposition is based on the work by \citet{Pathak2017_forward_model_intrinsic} and by \citet{Hafner2020_Dreamer}, who use a combined forward and inverse model to generate latent representations in a self-supervised manner. 
The bidirectional model operates on a latent space, generated by an abstraction function. 
By simultaneously performing inverse and backward prediction in the latent space, the abstraction function gains two important features: First, the latent space is de-noised because only the predictable parts of the latent observation determine the forward and backward prediction. Second, the method is self-supervised and, therefore, provides a more stable feedback signal than the comparatively weak reward signal.
This alleviates the prediction error and strengthens the overall learning signal. 

\section{Conclusion}
\label{sec:conclusion}
We provide an overview of the cognitive foundations of hierarchical problem-solving and how these are implemented in current hierarchical reinforcement learning architectures.
Herein, we consider few-shot problem-solving as the ultimate ability of intelligent beings that learn to solve causally non-trivial problems with as few trials as possible. 
As our main research question, we ask for the computational prerequisites and mechanisms to enable problem-solving for artificial computational agents on the level of intelligent animals. 
We further pose the challenge to integrate the different mechanisms. 

We appreciate that the cognitive theories we summarise are often contentious and controversial, as indicated in the \nameref{sec:cognitive_background} section. 
However, in our discussion, we show that the particular combination of the described theories have the potential to yield algorithmic synergies. 
In other words, we reason that if a particular mechanism is useful for computational architectures, then it is likely to be beneficial for biological cognitive systems as well, and vice versa. 
In this way, cognitive sciences and artificial intelligence inspire each other, and they will hopefully converge to insights that are relevant for both fields and useful for the future of AI and humanity.

\section*{Acknowledgements}
The authors acknowledge funding by the DFG (projects IDEAS, LeCAREbot, TRR169, SPP 2134, RTG 1808, EXC 2064/1) and the Humboldt Foundation and Max Planck Research School IMPRS-IS.


\balance 
\begin{sloppypar}
\setlength\bibitemsep{0pt}
\printbibliography[segment=\therefsegment,check=onlynew]
\end{sloppypar}

\newrefsegment



\renewcommand{\infobox}[1]
{\refstepcounter{infobox}\label{#1}}
\renewcommand{\table}[1]
{\refstepcounter{table}\label{#1}}
\infobox{box:appendix:glossary}
\infobox{box:appendix:meta_RL-survey}
\infobox{box:appendix:hrl_review}
\infobox{box:appendix:compositionality-example}
\infobox{box:appendix:automatization}
\infobox{box:appendix:EST}
\table{tab:hrl_properties}

\end{document}